\documentclass[letterpaper, 10pt, conference]{ieeeconf}
\IEEEoverridecommandlockouts
\usepackage{cite}
\usepackage{amsmath,amssymb,amsfonts}
\usepackage{booktabs}
\usepackage{textcomp}
\usepackage{xcolor}
\usepackage{graphicx}
\usepackage{threeparttable}
\usepackage{flushend}

\usepackage[labelfont=bf,labelsep=period,font=footnotesize]{caption}
\usepackage{subcaption}
\usepackage{url}
\usepackage{hyperref}
\hypersetup{hidelinks}
\begin{document}

\title{\LARGE \bf SABER: Learning Attention-based Semantic Affordance for \\ Legged Locomotion}

\author{Hari Prasanth Palanivelu$^{1}$, Samuel Sze$^{1}$, Kennard Garrison Johannes$^{2}$,\\
Albertus Hendrawan Adiwahono$^{1}$, and Meng Yee (Michael) Chuah$^{1}$\\[0.6ex]
{\normalsize Project page: \href{https://mcx-lab.github.io/saber-page/}{\texttt{mcx-lab.github.io/saber-page}}}%
\thanks{$^{1}$Authors are with the Institute of Advanced Intelligence and Computing
(IAIC), Agency for Science, Technology and Research (A*STAR), Singapore.}%
\thanks{$^{2}$Nanyang Technological University,
Singapore.}%
\thanks{This research is supported by the Home Team Science and Technology Agency (HTX).}%
}

\maketitle

\begin{abstract}
Perceptive legged locomotion has advanced rapidly by integrating terrain
\emph{geometry} into learned policies, yet the
integration of terrain \emph{meaning} remains sparse: a pipe, a patch of
grass, or a fragile box may be geometrically traversable while being inappropriate for contact.
In industrial environments, where legged robots increasingly operate, a single misplaced step can damage fragile equipment, destabilize the robot, or endanger the site.
To address this, we introduce SABER, a planner-free reinforcement-learning policy that jointly reasons about terrain geometry and semantic contact permission. The policy consumes a unified terrain-affordance map, where each cell encodes local 3D geometry and a \emph{semantic contact cost}. 
We augment cross-attention with a learned, signed semantic bias: an additive term on the attention logits, gated by the contact cost, that reweights flagged cells by their distance from the nearest foot. A hazard therefore reshapes attention where it can still affect the next foothold, and its influence fades where it cannot.
The resulting policy selects footholds on permitted support and keeps the leg clear of forbidden regions throughout the swing phase.
We perform a systematic ablation that isolates the contribution of each architectural component; removing the semantic bias alone increases forbidden contacts by 55\% while velocity tracking is unchanged.
We validate the policy on a Unitree B2, demonstrating sim-to-real semantic contact selection across indoor and outdoor environments and four semantic obstacle classes. 
\end{abstract}

\section{Introduction}
Learning-based legged locomotion has progressed from blind traversal of rough
terrain~\cite{hwangbo2019learning,lee2020learning,kumar2021rma} to perceptive
controllers that anticipate terrain before contact
occurs~\cite{miki2022wild,agarwal2023legged}, and more recently to agile
behaviors over stairs, gaps, stepping stones, and parkour
obstacles~\cite{zhuang2023robot,cheng2024extremeparkour,hoeller2024anymalparkour,kang2026agile}.
This progress has been enabled by several advances, including massively
parallel simulation~\cite{rudin2022learning} and learned terrain
representations that expose task-relevant geometry to the policy
~\cite{miki2022wild,agarwal2023legged,yang2021locotransformer,he2025ame}.
Ha et al. survey this maturity in geometry-aware locomotion~\cite{ha2025learning}. 
More recently, Frey et al.~\cite{frey2026advances} term the combination of
semantic understanding and careful, environment-conditioned foot placement
\emph{dexterous semantic locomotion}, framing it as a key open frontier.

Geometry describes where support exists, but not whether contact is appropriate. 
A pipe or fragile object may be physically usable as a foothold yet should not be stepped on, while grass or a puddle may be geometrically flat but undesirable. 
Geometric traversability and semantic contact permission are therefore distinct. Current systems resolve the distinction above the locomotion controller, with semantic traversability estimates and cost maps feeding navigation~\cite{wellhausen2019where,frey2023fast,kim2024semantic,roth2024viplanner},
while the policy beneath stays geometric~\cite{frey2026advances}. 
Bringing permission into the controller remains rare: semantics has selected speed and gait~\cite{yang2023semantics}, or routed footholds around clutter via a separate search~\cite{liang2026semloco}.

A geometry-only policy therefore faces an observation ambiguity: two scenes with identical geometry, robot state, and commands can require different contact decisions, because the same surface is permitted in one and forbidden in the other. 
Because the distinction is absent from the observation, no reward shaping or additional training can recover it.
We address this by bringing semantics into the low-level policy itself (Fig.~\ref{fig:teaser}): geometry and contact permission share one terrain-affordance map, and a learned, signed, distance-structured semantic bias on the attention logits tells the policy where flagged terrain matters relative to its feet. 
Our contributions are threefold:
\begin{figure}[!t]
  \centering
  \includegraphics[width=\columnwidth]{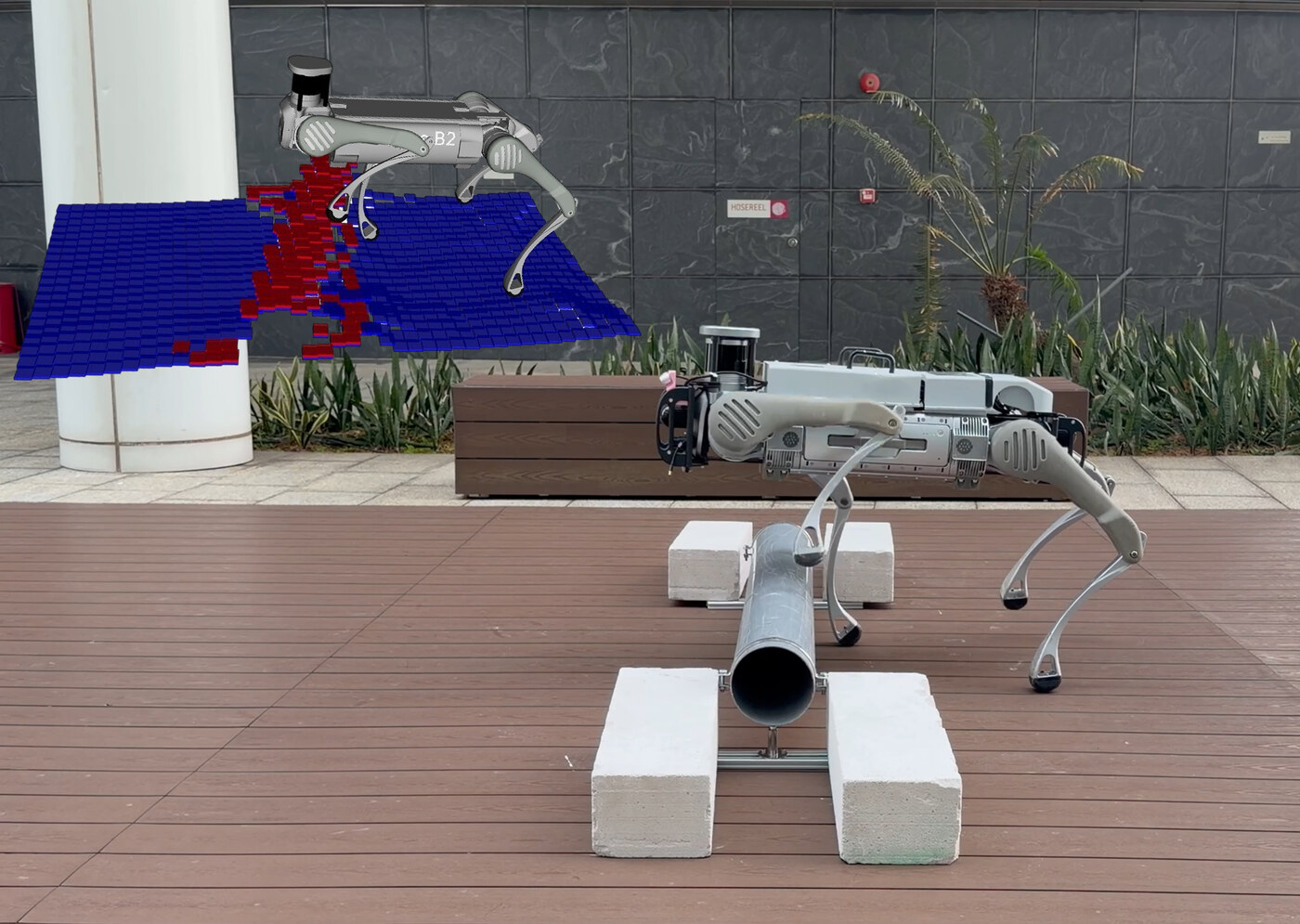}
  \caption{\textbf{Semantic contact selection on hardware.} The Unitree B2
    crosses a flagged pipe without touching it (right); in the corresponding
    terrain-affordance map (left), the pipe's cells carry contact cost $r=1$
    (red) while geometrically similar unflagged terrain remains available
    for support.}
  \label{fig:teaser}
\end{figure}

\begin{itemize}
    \item \textbf{Semantic contact-affordance formulation:}
    a unified per-cell $[x,y,z,r]$ terrain-affordance map that enables the low-level
    controller to separate where the robot \emph{can} step from where it
    \emph{should}, at foothold resolution.

    \item \textbf{Structured semantic attention:}
    a multi-head cross-attention terrain encoder whose logits carry a learned, signed,
    distance-structured bias, gated by the contact cost, so that each head
    weighs flagged terrain by its distance from the nearest foot.

    \item \textbf{Sim-to-real deployment and validation:}
    training terrains that decorrelate geometry from contact permission, and deployment on a
    Unitree B2 indoors and outdoors across four obstacle classes with
    off-the-shelf perception in place of simulator ground truth.
\end{itemize}

\section{Related Work}

\subsection{Learning-Based Perceptive Locomotion}

Learned locomotion controllers see the terrain through a deliberately narrow
channel. The policy receives a robot-centric height field sampled from an
elevation map~\cite{miki2022wild,he2025ame} or an egocentric depth
image~\cite{agarwal2023legged,cheng2024extremeparkour} and is trained in
simulation with privileged terrain information~\cite{lee2020learning}.  
Geometric exteroception is favored over color
images because it is what simulation reproduces with fidelity, and the
sim-to-real gap for rendered appearance remains
unresolved~\cite{frey2026advances}.
The price is that everything the geometric representations omit must be recovered after
contact. 
Miki et al. list material and texture among the properties the map
discards; deep snow, soft vegetation, and foam blocks all appear in the map
as solid surfaces, and the belief encoder revises its estimate only once a
foot has sunk in~\cite{miki2022wild}. 
The correction is robust, but it comes
after the step that a semantic input would have prevented. 
Later agile controllers sharpened the geometric reading rather than widened it. Extreme Parkour penalizes foot contacts within $5\,\mathrm{cm}$ of a terrain edge~\cite{cheng2024extremeparkour}, and AME's attention concentrates on the cells the robot is about to step on~\cite{he2025ame}. In each case the foot goes where the geometry allows. 
The omission is acknowledged from inside this line of work: PIE, which
clears gaps of three body lengths from egocentric depth, names the lack
of RGB semantics among its stated limitations~\cite{luo2024pie}.

\subsection{Attention Mechanisms in Locomotion Control}
 
Spatial attention over exteroceptive tokens is increasingly prevalent in locomotion
policies. LocoTransformer attends jointly over depth-image patches and a
proprioceptive token~\cite{yang2021locotransformer}; PIE fuses depth features
with the proprioceptive history in a shared transformer
encoder~\cite{luo2024pie}; and AME turns each cell of a local height map into
a token, attends over the full set with a single proprioceptive query, and
reads the attention weights as the footholds the policy is about to
use~\cite{he2025ame}. AME-2 adds a global map summary to the query and a
per-cell uncertainty channel~\cite{zhang2026ame2}, and TAGA makes the
selection itself learnable by predicting a gaze point and cropping the height
scan around it before attending~\cite{li2026taga}.
 
In all of these the attention score is the dot product of query and key.
Structure enters through what becomes a token, or through a fixed mask, as
in the Body Transformer, where the kinematic tree decides which body tokens
may exchange information~\cite{sferrazza2025bodytransformer}. 
A mask is a bias fixed at $-\infty$. 
The intermediate case, a learned additive term that reweights
scores without removing tokens, is standard in sequence models, where it
encodes relative position~\cite{liu2021swin,raffel2020exploring,press2021train}.
To our knowledge, no locomotion policy has placed a learned, signed,
spatially structured bias on its attention scores, and none has used one
to carry semantics. 
That is the mechanism this paper introduces, with the
distance to the nearest foot in place of sequence position and the contact
cost as the gate.

\subsection{Semantic Integration with Legged Locomotion}
Semantics usually acts one level above the controller. 
Traversability is learned from proprioceptive experience~\cite{wellhausen2019where, frey2023fast} or human video~\cite{kim2024semantic}, and ViPlanner turns a semantic image into per-class costs and plans a path for a separate locomotion policy~\cite{roth2024viplanner}. 
The path routes around a forbidden region, but the controller beneath stays geometric~\cite{frey2026advances}, so a foot that must cross one is placed without knowing it is forbidden. 
When semantics reaches the controller, it selects a behavior, not a contact: Yang et al.\ map an RGB embedding to speed and gait~\cite{yang2023semantics}, and LocoVLM retrieves a locomotion style from a vision-language description of the scene~\cite{nahrendra2026locovlm}, leaving foot placement to the underlying controller.

At the foothold level, semantic information is either derived from geometry or acted on outside the policy.
ViNL penalizes stepping on floor clutter inside a learned policy, but the clutter is seen through depth, so the policy avoids shapes, not classes~\cite{kareer2023vinl}. 
QuadPiPS segments depth into steppability labels that gate its foothold search~\cite{asselmeier2026quadpips}; the labels are defined by shape, so they sharpen geometry rather than add a permission it lacks. 
Closest to ours is SemLoco~\cite{liang2026semloco}, which trains a policy to avoid low-lying clutter from a per-cell fragility cost. 
The policy observes the cost, but a Raibert-heuristic grid search picks the foothold for it to track, so the decision is made outside the network; and because its hazards protrude from a flat floor, an elevation-only ablation matches its success rate at low clutter density.
No existing policy consumes a semantic contact cost at foothold resolution
and chooses the foothold itself; ours does.
It is planner-free: the cost enters the attention that selects the foothold
rather than a separate search.
It is trained on terrains that decorrelate geometry from permission: raised supports forbidden only by their cost, and flat ground forbidden with no geometric sign. 
And its learned prior is inspectable: eight gains and eight radial profiles expose which heads use the cost and at what range.

\begin{figure*}[!t]
  \centering
  \includegraphics[width=\textwidth]{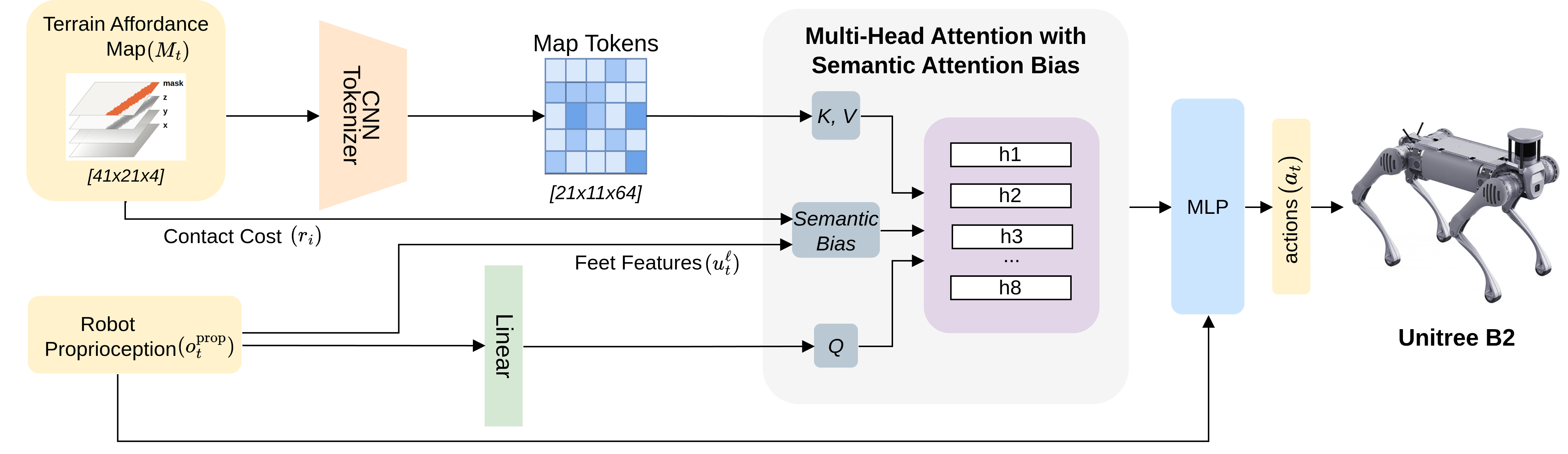}
    \caption{\textbf{Overview of the proposed framework.} A CNN tokenizes the terrain-affordance map and a single proprioceptive query attends over the tokens; the contact cost enters both as a CNN input channel and as a distance-structured semantic bias on the attention logits.
    }  
  \label{fig:framework}
\end{figure*}

\section{Method}
\label{sec:method}

\subsection{Overview}
\label{sec:overview}

We formulate locomotion under contact restrictions as velocity tracking in a
partially observable Markov decision process and solve it with model-free
reinforcement learning. At each control step, the policy receives an
observation $o_t$ and outputs a 12-dimensional action
$a_t \sim \pi_{\theta}(a_t \mid o_t)$ that parameterizes joint position
targets around a nominal posture,
$q_t^{\mathrm{des}} = q^{\mathrm{nom}} + s_a \odot a_t$
with a fixed action scale $s_a$, tracked by joint-level PD control.
Policy and value function are optimized with PPO
\cite{schulman2017ppo} in massively parallel simulation using a
velocity-tracking objective augmented with penalties for contact on
forbidden terrain (Sec.~\ref{sec:training}).
We use an asymmetric actor-critic formulation: the actor observes only
quantities available on the real robot, while the critic additionally
receives privileged simulator state.

Fig.~\ref{fig:framework} illustrates the proposed framework.
The local environment is represented by a yaw-aligned terrain-affordance
map $M_t$, where each cell
$m_{t,i}=[x_{t,i},y_{t,i},z_{t,i},r_{t,i}]$ contains local 3D geometry and
a semantic contact cost $r_{t,i}$, with $r=0$ for steppable terrain and
$r=1$ for terrain that must not be contacted; we call cells with $r=1$
\emph{flagged}. A convolutional tokenizer converts the
map into terrain tokens that serve as the keys and values of a
cross-attention layer, while the proprioceptive observation, including
per-foot kinematics, forms a single query:
\begin{equation}
    Q = f_q(o_t^{\mathrm{prop}}), \qquad
    K,V = f_{kv}(M_t^{\mathrm{height}},M_t^{\mathrm{semantic}})
\end{equation}

The contact cost reaches the attention both through the terrain tokens
and through a structured additive bias on the attention logits
(Sec.~\ref{sec:bias}). The attended terrain context is combined with
proprioception and decoded into joint targets. The policy is trained
end-to-end and requires no footstep planner, reference trajectory, or
precomputed footholds.

\subsection{Terrain-Affordance Map and Tokenization}
\label{sec:map}

The map $M_t$ is an $L \times W$ grid in a yaw-aligned frame attached to
the base, covering $2.0\,\mathrm{m} \times 1.0\,\mathrm{m}$ at
$0.05\,\mathrm{m}$ resolution ($41 \times 21$ cells) and centered
$0.2\,\mathrm{m}$ ahead of the base so that most of the map lies in the forward
walking direction.
The geometric channels, $M_t^{\mathrm{height}}$, of a cell hold the local $x$, $y$, $z$ of a downward
ray hit; heights are base-relative and clipped to $[-1.2,0]\,\mathrm{m}$, with ray misses assigned $-1.2\,\mathrm{m}$.
The semantic cost channel, $M_t^{\mathrm{semantic}}$, holds the contact cost $r$ of each cell, and it is produced
by a second ray caster that observes only flagged geometry and shares the
frame, pattern, and ray ordering of the geometric scan, so the two channels
are aligned cell by cell by construction. 
The contact cost is binary in this work, so the channel is a contact mask; the interface can be extended to continuous costs without architectural change. 
 In simulation, the cost derives from ground-truth object identity; at deployment, it derives from projected segmentation fused into a semantic elevation map (Sec.~\ref{sec:deployment_pipeline}).

A lightweight CNN tokenizes the four-channel map. It consists of a
$5\times5$ convolution with stride 2 and padding 2
(4$\rightarrow$16 features), followed by a $3\times3$ convolution with
stride 1 and padding 1 (16$\rightarrow$64 features); each layer is followed
by ReLU and batch normalization. The resulting $21\times11$ grid contains
$N=231$ terrain tokens $k_{t,i}\in\mathbb{R}^{64}$, which serve as the keys
and values of the attention layer. Each token has a $9\times9$-cell
($0.45\,\mathrm{m}$) receptive field and therefore represents a local
neighborhood rather than a single height sample.
All four channels enter the convolution jointly, so a token can encode conjunctions such as a height discontinuity that is also
flagged. 
We add no positional encoding and do not concatenate coordinates
to the tokens as in \cite{he2025ame}: the cell coordinates $x$ and $y$ are input features, which places every token in the egocentric frame and lets attention reason about where a cell is without an auxiliary embedding.

Fusing geometry and semantics at the CNN input provides the necessary information but does not guarantee its effective use (Sec. IV-A.1), which motivates the direct route introduced next.

\subsection{Structured Semantic Attention Bias}
\label{sec:bias}

The terrain tokens are passed through a single cross-attention layer with
eight heads. Each head projects the $64$-dimensional tokens into its own
$8$-dimensional key and value subspace, so $d_h = 8$, and the eight attended
outputs are concatenated and linearly mixed back to $64$ dimensions.
The query is a linear projection of the proprioceptive observation,
$q_t = W_q\, o_t^{\mathrm{prop}}$. Among its entries, $o_t^{\mathrm{prop}}$
carries, for every foot $\ell \in \{\mathrm{FR}, \mathrm{FL}, \mathrm{RR}, \mathrm{RL}\}$,
the position and velocity of that foot relative to the base,
\begin{equation}
    u_t^{\ell} = \left[\, p_t^{\ell},\; \dot{p}_t^{\ell} \,\right],
    \label{eq:limb_features}
\end{equation}
expressed in the same yaw-aligned frame as the map and obtained from the
joint encoders by forward kinematics.

Through the tokens, the contact cost reaches a head only after sharing every convolution filter and the $64 \to 8$ key projection with the geometric channels. Its influence on the attention scores may therefore be attenuated, so we provide a second, direct route: an additive term on the pre-softmax logits. Additive logit
biases are an established way of injecting structured side information
into attention without masking; sequence transformers use them to encode
token position~\cite{liu2021swin, raffel2020exploring, press2021train};
we adapt this idea to spatial contact semantics. Each head
$h$ computes
\begin{equation}
\mathrm{Attention}(Q_h, K_h, V_h)
= \mathrm{softmax}\!\left(
\frac{Q_h K_h^{\top}}{\sqrt{d_h}} + B_h
\right) V_h ,
\label{eq:biased_score}
\end{equation}
where $Q_h$, $K_h$ and $V_h$ are the projected query, key and value
matrices, the softmax normalizes over the $N$ terrain tokens, and
$B_h \in \mathbb{R}^{1 \times N}$ is the row of bias values,
\begin{equation}
    B_{t,h,i}
    = \underbrace{\beta_h}_{\text{per-head gain}}
    \cdot
    \underbrace{\rho_h\!\left(d_{t,i}\right)}_{\text{radial profile}}
    \cdot
    \underbrace{\bar{r}_{t,i}}_{\text{contact cost}} .
    \label{eq:reach_bias}
\end{equation}

The gain $\beta_h$ sets how strongly head $h$ responds to flagged terrain.
We leave it signed and unconstrained rather than enforcing suppression,
allowing learning to determine whether a head suppresses, emphasizes, or
effectively ignores flagged terrain (Sec.~\ref{sec:attention_analysis}). With $\beta_h$ initialized to zero, the layer
reduces exactly to standard attention at the start of training.

The radial profile $\rho_h$ decides where flagged terrain matters. A flagged cell one stride ahead of a foot may directly affect its next contact, whereas an identical distant cell is generally less immediately relevant. Although there is a
single query, we give the bias a foot-centric structure by assigning each
cell to its nearest foot in the ground plane,
\begin{equation}
    d_{t,i} = \min_{\ell}\,
    \bigl\lVert x_{t,i} - p_{t,xy}^{\ell} \bigr\rVert,
    \label{eq:nearest_foot}
\end{equation}
with $x_{t,i}$ the cell's in-plane centre in the map frame. This is a hard
partition with no radius cap. $\rho_h$ is a per-head piecewise-linear
function of $d$ with $K = 6$ control points spaced evenly from zero to the
map diagonal, which upper-bounds any foot-to-cell distance. All control
points start at one, so each head begins distance-independent and learns
its own range; a head that should ignore distant cells can flatten its
profile there, whereas a hand-picked cutoff radius would duplicate that
role with a constant to tune. Because $\rho_h$ is also unconstrained, the
sign of a head's response is carried by the product $\beta_h\,\rho_h(d)$
rather than by the gain alone.

The token-level cost $\bar{r}_{t,i}$ is the cell-level cost $r$ reduced to
token resolution by max pooling, so that thin flagged objects are not averaged
away. It gates the bias to flagged terrain, while $\rho_h(d_{t,i})$
determines how strongly individual flagged tokens are modulated. When no
terrain is flagged, $B_h=0$ and the layer reduces exactly to standard
attention. The complete mechanism comprises only eight gains and eight
six-point profiles, making its learned behavior directly inspectable
(Sec.~\ref{sec:attention_analysis}).

\subsection{Training Pipeline}
\label{sec:training}

\begin{figure*}[!t]
  \centering
  \includegraphics[width=\textwidth]{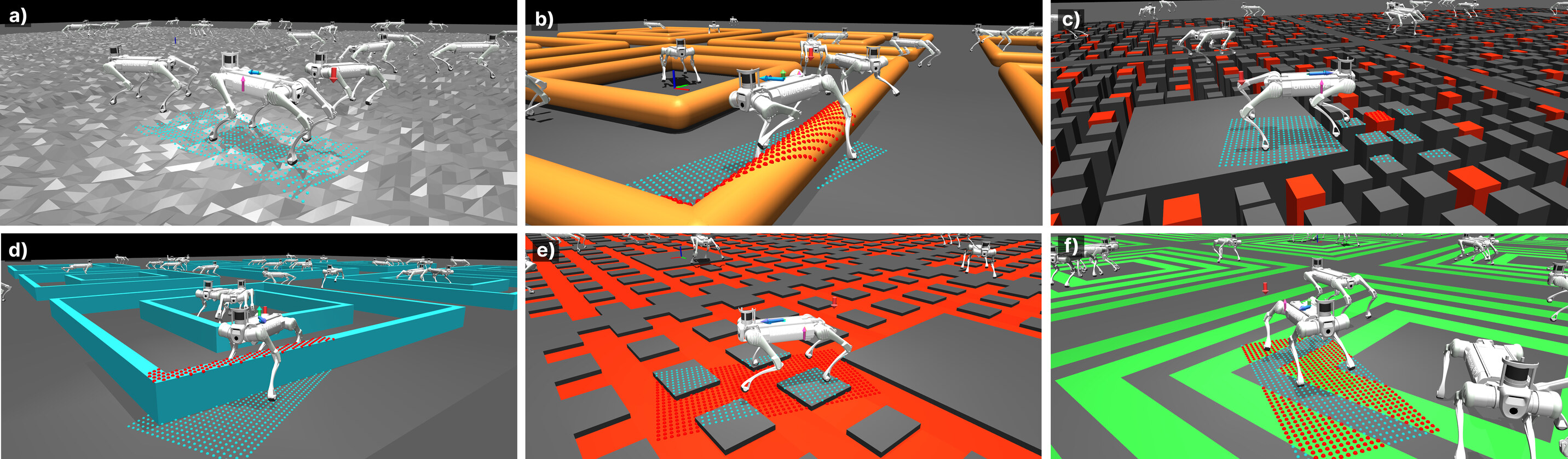}
  \caption{\textbf{Training terrain suite.}
    a) rough terrain;
    b) pipes;
    c) semantic stepping stones;
    d) rails;
    e) lava tiles;
    f) concentric grass rings.
    Colored terrain regions are flagged.
    In the terrain-affordance map, red points denote flagged cells and blue
    points denote unflagged cells.}
  \label{fig:terrains}
\end{figure*}

\begin{figure}[t]
  \centering
  \includegraphics[width=\columnwidth]{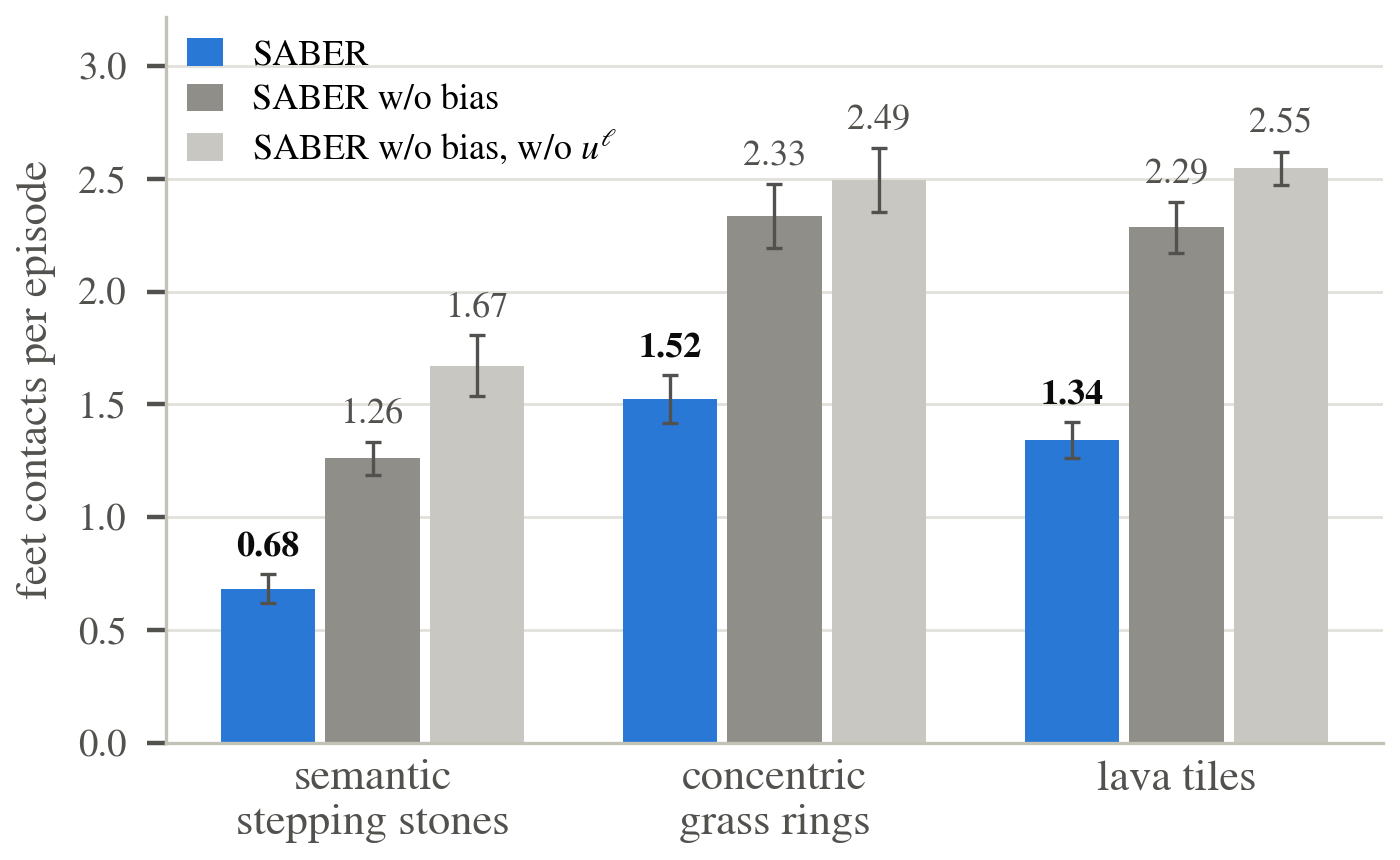}
  \caption{\textbf{Feet contacts per episode on the three terrains without a geometric cue.} On the stepping stones, the concentric grass rings, and the lava tiles, flagged and unflagged ground have identical geometry, so only the cost channel can tell them apart.}
   \label{fig:feet_contacts_bar}
\end{figure}

\subsubsection{Observations}
The actor observes
\begin{equation}
    o_t = \bigl[\,\omega_t,\; g_t,\; q_t,\; \dot{q}_t,\; a_{t-1},\;
    c_t,\;  u_t^{\ell},\; M_t\,\bigr],
\end{equation}
where $\omega_t$ is the base angular velocity, $g_t$ the projected gravity
vector, $q_t$ and $\dot{q}_t$ the joint positions and velocities, $a_{t-1}$
the previous action, $c_t$ the velocity command, $u_t^{\ell}$ the per-foot features, Eq.~\eqref{eq:limb_features}, and $M_t$ the
terrain-affordance map.

The critic augments this observation with privileged base linear velocity
and per-foot terrain clearance, air time, contact state, and contact force.
The critic shares the map tokenizer, the cross-attention layer, and the attention bias of Sec.~\ref{sec:bias} with the
 actor, but forms its own query from the critic observation $o^{c}_t$ and has its own value head, so
 the value loss trains the shared encoder jointly with the policy loss.

\subsubsection{Rewards}
The reward is a weighted sum of the terms in Table~\ref{tab:rewards},
where $\mathbb{I}(\cdot)$ denotes the indicator function and
$\mathcal{C}$ and $\mathcal{T}$ denote feet in contact and at touchdown,
respectively. The task objective is augmented with gait-shaping and
regularization terms, together with penalties for foot and shank contact
with flagged terrain. The upright reward is defined relative to the local
terrain normal, and gait terms are active only under nonzero motion
commands; peak swing clearance is evaluated at touchdown. For the joint-posture term, $\sigma_j = 0.05\,\mathrm{rad}$ for the hip and thigh joints and $\sigma_j = 0.1\,\mathrm{rad}$ for the calf joints while
standing; during motion, these tolerances are scaled by a factor of $6$. Episodes
terminate on thigh, trunk, or head terrain contact or when the base tilt
exceeds $70^\circ$.

\begin{table}[t]
\centering
\caption{\textbf{Reward terms and weights.}}
\label{tab:rewards}

\sbox0{%
\footnotesize
\setlength{\tabcolsep}{4pt}
\begin{tabular}{@{}lcr@{}}
\toprule
Reward & Expression $r_i$ & Weight $w_i$ \\
\midrule

\multicolumn{3}{@{}l}{\emph{\textbf{Task}}} \\
Lin.-velocity tracking
& $\exp(-4\,\lVert v^{\mathrm{cmd}}-v\rVert^{2})$
& $+3.5$ \\

Ang.-velocity tracking
& $\exp(-2\,\lVert\omega^{\mathrm{cmd}}-\omega\rVert^{2})$
& $+2.0$ \\

Upright posture
& $\exp(-5\sin^{2}\theta_n)$
& $+1.0$ \\

Joint posture
& $\exp\!\left(-\frac{1}{12}\sum_j(\Delta q_j/\sigma_j)^2\right)$
& $+1.0$ \\

Termination
& $n^{\mathrm{termination}}$
& $-100$ \\

\midrule
\multicolumn{3}{@{}l}{\emph{\textbf{Gait}}} \\

Feet air time
& $\sum_\ell
   \mathbb{I}\!\left(0.05<t_{\mathrm{air}}^\ell<0.5\right)$
& $+0.25$ \\

Foot clearance
& $\sum_\ell
   |h^\ell-0.15|\,\lVert\dot p^\ell_{xy}\rVert$
& $-2.0$ \\

Foot swing height
& $\sum_{\ell\in\mathcal{T}}
   (h^\ell_{\mathrm{peak}}/0.15-1)^2$
& $-0.25$ \\

Foot slip
& $\sum_{\ell\in\mathcal{C}}
   \lVert\dot p^\ell_{xy}\rVert^2$
& $-0.1$ \\

Soft landing
& $\sum_{\ell\in\mathcal{T}}\lVert F^\ell\rVert$
& $-10^{-5}$ \\

\midrule
\multicolumn{3}{@{}l}{\emph{\textbf{Regularization}}} \\

Joint position limits
& $\sum_j
   \max\!\left(
   |q_j-\bar q_j|-0.9\,q_j^{\mathrm{lim}},\,0
   \right)$
& $-1.0$ \\

Action rate
& $\lVert a_t-a_{t-1}\rVert^2$
& $-0.1$ \\

Self collision
& $n^{\mathrm{self}}$
& $-0.1$ \\

Shank contact
& $n^{\mathrm{shank}}$
& $-0.1$ \\

\midrule
\multicolumn{3}{@{}l}{\emph{\textbf{Semantic}}} \\

Foot--flagged contact
& $N^{\mathrm{foot}}$
& $-2.5$ \\

Shank--flagged contact
& $N^{\mathrm{shank}}$
& $-5.0$ \\

\bottomrule
\end{tabular}}%

\ifdim\wd0>\columnwidth
  \resizebox{\columnwidth}{!}{\usebox0}%
\else
  \usebox0%
\fi
\end{table}
\subsubsection{Terrains}
Training terrain (Fig.~\ref{fig:terrains}) mixes two kinds of hazard. Some
are visible in the geometry channels: flagged pipes and rails protrude from
the ground, and must not be touched. Others are visible only in the cost
channel: grass rings are flush with the ground, and lava lies only
$5\,\mathrm{cm}$ below the tile tops, so geometry reports steppable ground
in both cases. The stepping-stone field contains both kinds at once: the
gaps between the stones are a geometric hazard, while a difficulty-scaled
fraction of the stones is flagged, which geometry cannot distinguish from
the rest. Plain rough terrain contains nothing flagged and maintains
ordinary locomotion skill.

Two choices prevent the policy from solving all of this through geometry
alone. If every pipe were flagged, the policy could learn to avoid anything
that protrudes and ignore the cost channel entirely; we therefore
also train on pipes and rails that are geometrically identical to the
flagged ones but perfectly fine to step on, so that shape does not predict
permission. For the same reason, flagged and safe stepping stones share
identical geometry---only the cost channel tells them apart---and no
two flagged stones are adjacent, so a valid path always exists. 
Terrain difficulty follows the tracking-driven row curriculum of~\cite{rudin2022learning}.

\subsubsection{Training Setup}
\label{sec:schedule}

We train with mjlab~\cite{zakka2026mjlablightweightframeworkgpuaccelerated}
using PPO and its default velocity-tracking hyperparameters, except that
each epoch uses eight mini-batches to accommodate the attention encoder.
The actor and critic heads are $512\times256\times128$ ELU MLPs operating
on the concatenated attention output and proprioceptive features, and the
policy runs at $50\,\mathrm{Hz}$. Domain randomization follows the
mjlab velocity-tracking defaults.

Each policy is trained for 23k iterations with 4096 parallel environments.
Proprioceptive and foot-kinematic noise from
Table~\ref{tab:observation_noise} is active throughout training, while both map-height noise terms are introduced after 15k iterations and remain active for the
final 8k. The critic receives clean observations throughout. Training on a
single NVIDIA H200 takes approximately 34 hours, and the final checkpoint
is deployed without further adaptation.

\begin{table}[t]
\centering
\caption{\textbf{Actor observation noise.}}
\label{tab:observation_noise}
{\footnotesize
\setlength{\tabcolsep}{6pt}
\begin{tabular}{@{}lcc@{}}
\toprule
Observation & Noise & Unit \\
\midrule
Base angular velocity
& $\mathcal{U}(-0.2,\,0.2)$ & rad/s \\

Projected gravity
& $\mathcal{U}(-0.05,\,0.05)$ & -- \\

Joint position
& $\mathcal{U}(-0.01,\,0.01)$ & rad \\

Joint velocity
& $\mathcal{U}(-1.5,\,1.5)$ & rad/s \\

Foot position
& $\mathcal{U}(-0.03,\,0.03)$ & m \\

Foot velocity
& $\mathcal{U}(-0.25,\,0.25)$ & m/s \\

Map cell height
& $\mathcal{N}(0,\,0.03^2)$ & m \\

Map height offset
& $\mathcal{U}(-0.05,\,0.05)$ & m \\

\bottomrule
\end{tabular}}
\end{table}

\begin{figure*}[t]
  \centering
  \includegraphics[width=\textwidth]{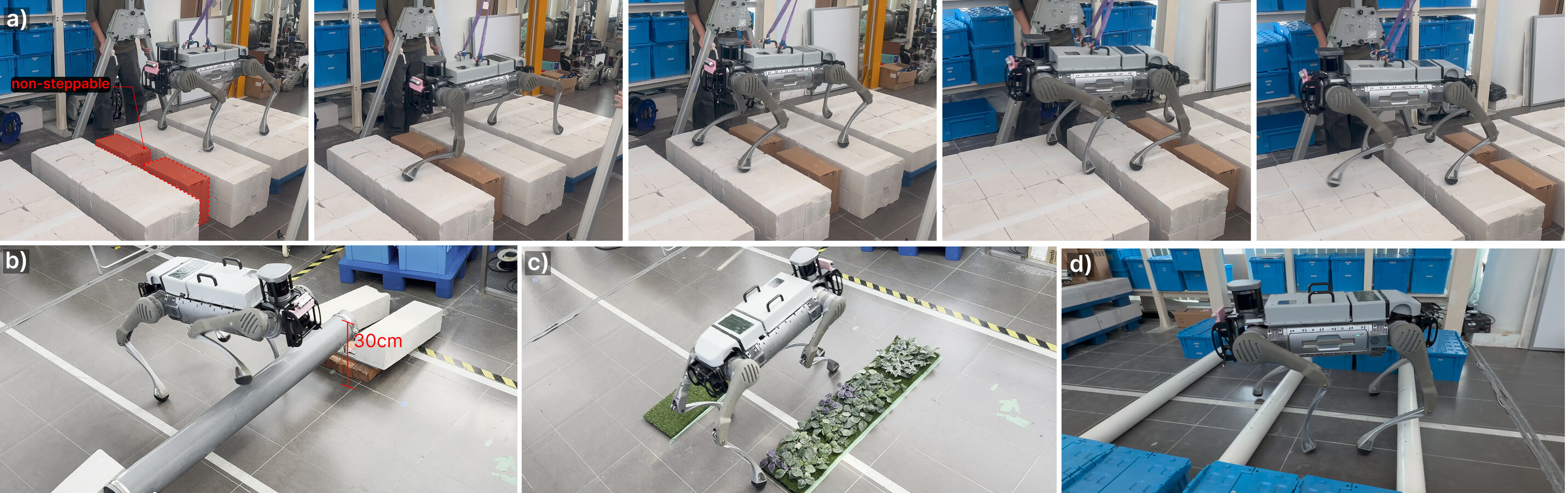}
  \caption{\textbf{Indoor scenes.} (a) Semantic stepping stones: the B2 steps on the concrete rails and keeps its feet off the flagged cardboard boxes. (b) B2 crossing a $30\,\mathrm{cm}$ pipe. (c) B2 crossing turf patches sideways. (d) Three low pipes, the tallest $11\,\mathrm{cm}$, laid parallel.}
   \label{fig:indoor_pipes_grass}
\end{figure*}

\section{Experiments and Results}

\subsection{Simulation Experiments}
\label{sec:sim_exps}

\subsubsection{Architecture Ablation}
\label{sec:ablation}

We designed six ablations against SABER described in Sec.~\ref{sec:method}, each altering one element while the actor and critic heads and everything in Sec.~\ref{sec:training}, including the seed, stay fixed, so that each comparison isolates that ingredient.
\begin{itemize}
  \item \emph{SABER w/o bias}: fixes $\beta_h = 0$, so the contact cost
    reaches the policy only through the tokens.
  \item \emph{SABER w/o bias, w/o $u^{\ell}$}: additionally removes
$u_t^{\ell}$ from the query, leaving the plain four-channel encoder.
  \item \emph{SABER, static bias, w/o $u^{\ell}$}: keeps that query and
restores only the gain, $B_{t,h,i} = \beta_h\,\bar{r}_{t,i}$, a uniform
prior on flagged cells with no distance structure.
  \item \emph{SABER w/o semantics, w/o bias, w/o $u^{\ell}$}: the
    plain encoder on the three geometric channels alone, with the cost
    channel and the two flagged-contact penalties removed. It is the floor against which
    every other row is read.
  \item \emph{SABER, 16 heads} and \emph{SABER, 32 heads}: the full model
    with the $64$ attention dimensions split into sixteen heads of
    dimension four or thirty-two heads of dimension two.
\end{itemize}

We report four quantities per episode: 
\emph{Contacts}, the number of distinct contact events between any foot and flagged geometry, our primary metric; 
\emph{Velocity error}, the planar tracking error $\lVert v^{\mathrm{cmd}}_{xy} - v_{xy}\rVert$ averaged over the episode;
\emph{Success}, the fraction of episodes that reach the time horizon without a termination; and
\emph{Distance}, the path length traveled. 

Each variant is evaluated separately on five terrain types using a
$6\times6$ grid at maximum training difficulty. For each terrain, 256 robots
are initialized at random positions and headings, each contributing one
$10\,\mathrm{s}$ episode. Commands are sampled from the training distribution
with a minimum linear speed of $0.3\,\mathrm{m/s}$ and a yaw rate of
$0.25\,\mathrm{rad/s}$ to exclude stationary trials. Each evaluation is
repeated over ten seeds that jointly resample the terrain layout, spawn
positions, and commands; observation noise is disabled. The terrains comprise
geometric hazards (flagged pipes and rails), semantic-only hazards (grass
rings and lava tiles), and stepping-stone fields combining both.
Table~\ref{tab:ablation} reports their equally weighted mean.

\begin{table}[t]
\centering
\caption{\textbf{Architecture ablation}, averaged over the five evaluation terrains
and ten evaluation repeats. Contacts are events per episode, velocity error
is in m/s, distance in m.}
\label{tab:ablation}
\sbox0{%
\scriptsize
\setlength{\tabcolsep}{3pt}%
\begin{threeparttable}
\begin{tabular}{@{}lcccc@{}}
\toprule
Variant & Contacts $\downarrow$ & Vel.\ err.\ $\downarrow$ & Success $\uparrow$ & Dist.\ $\uparrow$ \\
\midrule
\textbf{SABER}                      & \textbf{1.04} & 0.185          & \textbf{0.908} & 6.15 \\
SABER, 16 heads                     & 1.11          & \textbf{0.173} & 0.897          & \textbf{6.23} \\
SABER, 32 heads                     & 1.92          & 0.199          & 0.860          & 5.96 \\
SABER, static bias, w/o $u^{\ell}$  & 1.71          & 0.221          & 0.896          & 5.94 \\
SABER w/o bias                      & 1.61          & 0.185          & 0.905          & 6.18 \\
SABER w/o bias, w/o $u^{\ell}$      & 1.70          & 0.230          & 0.905          & 5.89 \\
SABER w/o semantics, w/o bias, w/o $u^{\ell}$ & 13.51 & 0.178\rlap{\tnote{\dag}} & 0.905 & 6.11 \\
\bottomrule
\end{tabular}
\begin{tablenotes}
\item[\dag] Not a merit: without the cost channel, the policy is never
 required to avoid flagged regions, so it tracks the commanded velocity
 unimpeded while stepping through them.
\end{tablenotes}
\end{threeparttable}}%
\ifdim\wd0>\columnwidth
  \resizebox{\columnwidth}{!}{\usebox0}%
\else
  \usebox0%
\fi
\end{table}

Removing the cost channel altogether raises contacts to $13.5$ per episode, thirteen times the full model, at the same success rate and distance: a policy that cannot see the cost simply walks through the flagged regions.
The distance-structured bias is what makes the policy selective.
\emph{SABER w/o bias} touches flagged geometry 55\% more, while velocity error, success, and distance stay unchanged.
Our model gains the most where geometry offers no cue: on the lava tiles, the grass rings, and the stepping stones, flagged and safe ground are geometrically indistinguishable (Fig.~\ref{fig:feet_contacts_bar}).
Finally, splitting the same $64$ dimensions across more heads trades one metric for another. Sixteen heads of dimension four give the lowest velocity error and the longest traverse in the table, but touch flagged geometry $7\%$ more often than eight heads. Thirty-two heads of dimension two is the weakest semantic variant, with $85\%$ more contacts than the full model and the lowest success rate. We conclude that a two-dimensional projection of the token embedding is too small to carry geometry and semantics at once, and that the per-head bias cannot compensate for keys that no longer carry what it reweights.

\subsubsection{Attention and Bias Analysis}
\label{sec:attention_analysis}

\begin{figure}[!t]
  \centering
  \includegraphics[width=\columnwidth]{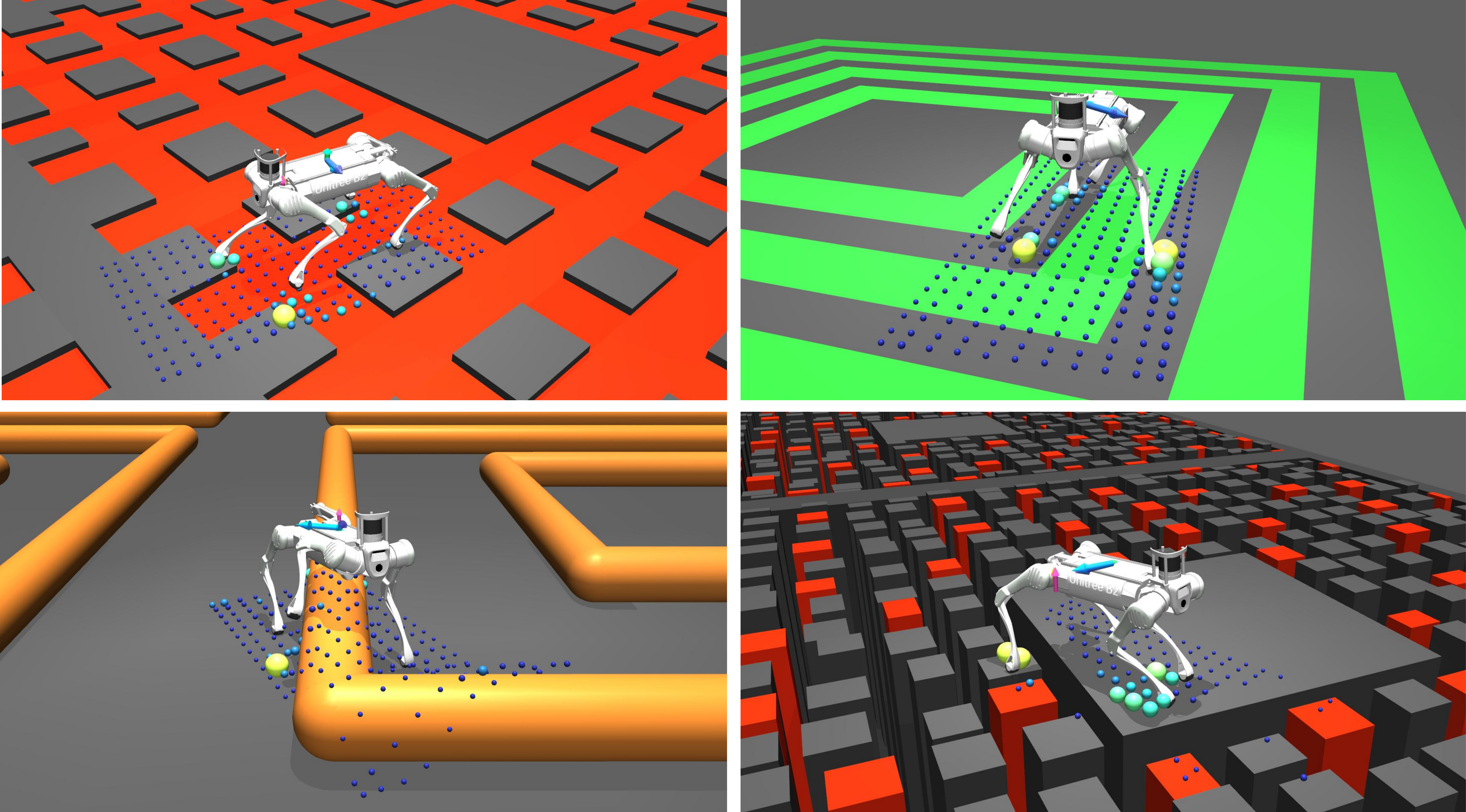}
      \caption{\textbf{Visualization of attention in mjlab}. Attention over the $N = 231$ terrain tokens, averaged over the eight heads, on four training terrains. Small blue markers are map cell centres in the yaw-aligned robot frame; sphere size and colour encode the attention weight a cell receives. Uniform attention would assign $1/231 = 0.004$ to every token.}   
  \label{fig:attention_viz}
\end{figure}

\begin{figure}[t]
  \centering
  \includegraphics[width=\columnwidth]{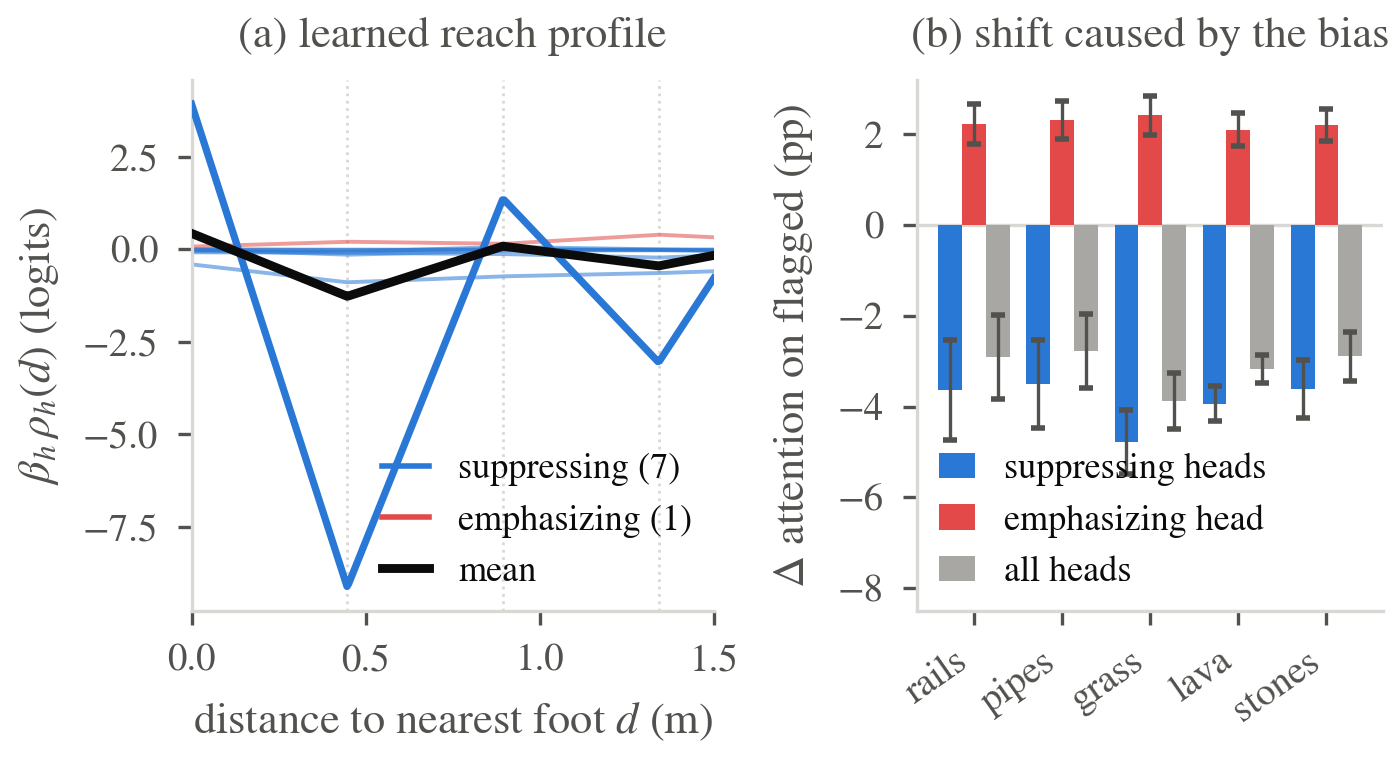}
  \caption{\textbf{Learned semantic attention bias.} (a) Effective bias
$\beta_h\,\rho_h(d)$ of each head against the distance $d$ to the nearest
foot. (b) Change in
attention mass on flagged cells caused by the bias, per terrain. Negative means the bias diverts attention away from
flagged cells.}
  \label{fig:bias_profile}
\end{figure}

Fig.~\ref{fig:bias_profile}(a) shows distinct behavior across heads. Six of the eight retain a small gain ($|\beta_h|<0.2$); of the other two, the stronger learns an effective bias that is slightly positive near the foot, reaches approximately $-9$ logits at $0.45\,\mathrm{m}$, and decays toward zero beyond $1\,\mathrm{m}$. Replaying identical observations with and without the learned bias ($\beta_h=0$) across 64 robots and 1200 steps per terrain shows that the bias removes about three percentage points of attention mass from flagged cells on every terrain, whether or not the hazard has a geometric signature. The effect is heterogeneous: some heads suppress flagged regions while one emphasizes them. Together with the static-bias ablation of Sec.~\ref{sec:ablation}, which does not reduce contacts, this indicates that the benefit comes from the learned foot-relative distance structure rather than from a generic per-head gain.

Fig.~\ref{fig:attention_viz} shows the head-averaged attention on four training terrains. It concentrates on the cells under the stance feet and on the cell where the swing foot is about to land, and flagged cells still receive a little attention rather than none: the bias is a prior, not a mask. The pattern is also phase dependent: early in a swing the attention is spread out and includes flagged cells; toward touchdown it collapses onto the landing cell, leaving almost none on flagged terrain at the moment that decides where the foot goes. The landing cell stands out before contact although the architecture contains neither a footstep planner nor a foothold target.

\begin{figure}[t]
  \centering
  \includegraphics[width=\columnwidth]{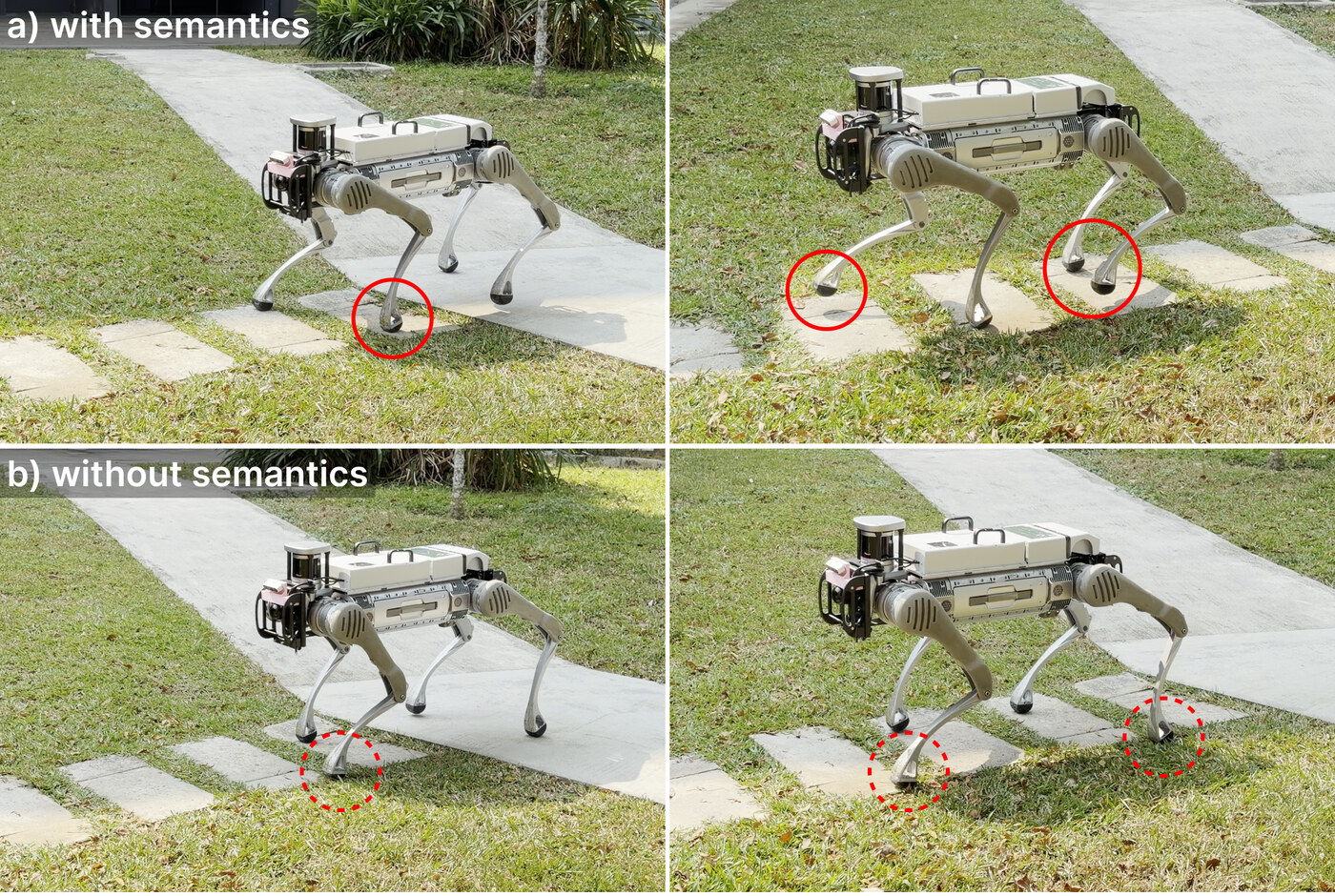}
  \caption{\textbf{Outdoor tiles with grass gaps}, the hardware counterpart of the lava tiles. (a) With semantics active, the B2 places its feet on the tiles (solid circles). (b) Without semantics, the same policy steps onto the grass (dashed circles).}
   \label{fig:outdoor_lava_tiles}
\end{figure}

\begin{figure}[t]
  \centering
  \includegraphics[width=\columnwidth]{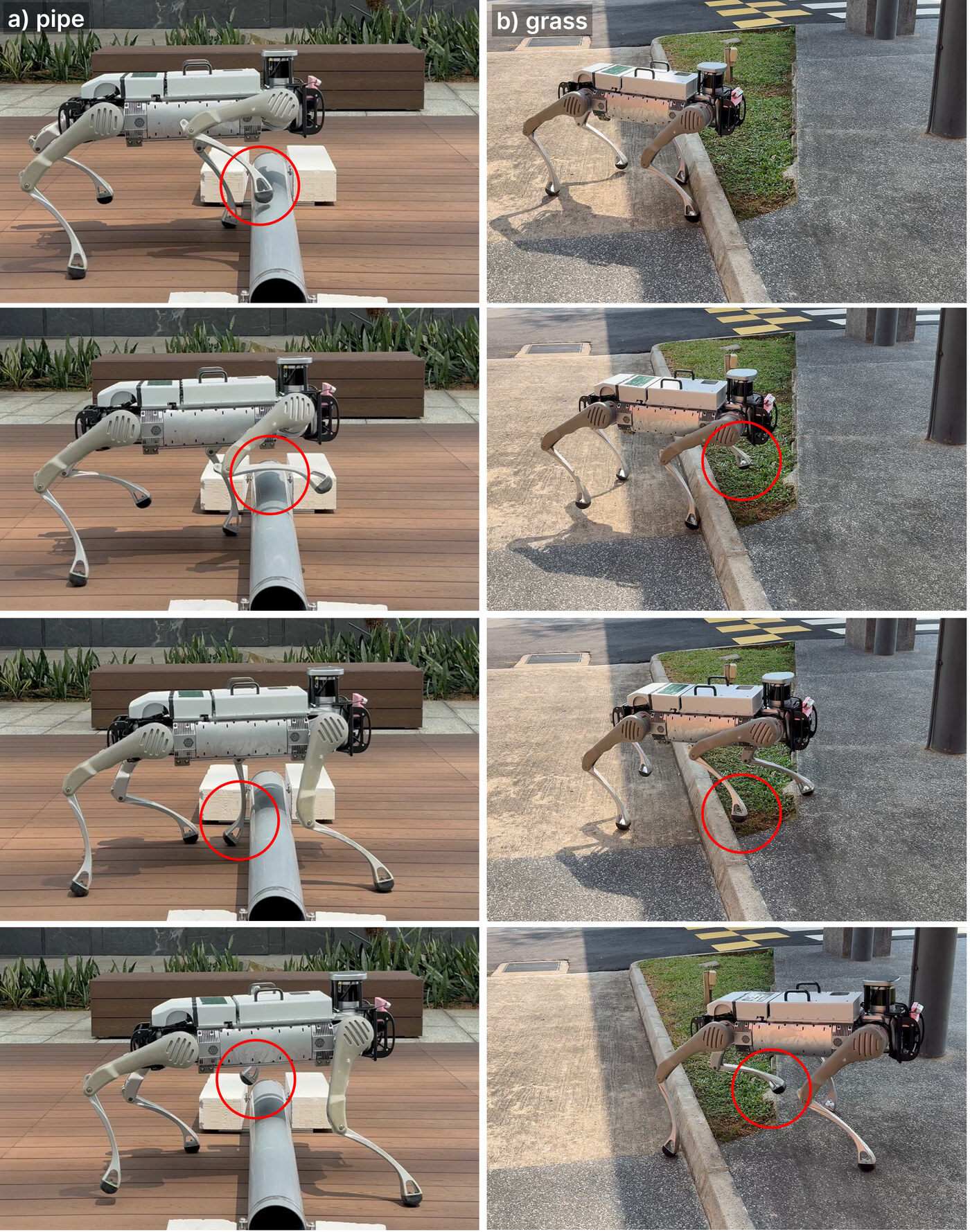}
  \caption{\textbf{Outdoor pipe and grass.} (a) Crossing a pipe on a timber deck. (b) Walking along a strip of grass beside a pavement. The circle marks the foot nearest the flagged object, which passes over the pipe and stays on the pavement.}
   \label{fig:outdoor_pipe_and_grass}
\end{figure}

\subsection{Hardware Deployment}
\label{sec:hardware_deployment}

\subsubsection{Deployment Pipeline}
\label{sec:deployment_pipeline}
We deploy our policy on the quadruped robot platform Unitree B2, with all perception, mapping, and policy inference running onboard.
The B2 is equipped with a RoboSense Helios-32 LiDAR with Direct LiDAR Odometry
(DLO)~\cite{dlo} for base pose estimation, forward- and rear-facing
RoboSense Airy LiDARs for elevation mapping, and a forward-facing Intel
RealSense D435i RGB camera. Forbidden support regions are segmented with a
YOLO26s-seg model~\cite{jocher2026yolo26}, fine-tuned on more than
6{,}000 annotated images. The policy consumes contact permission rather than
class identity; in our experiments, grass, vegetation, pipes, and cardboard
boxes are flagged and collapsed into a single cost channel.
Geometry and segmentation scores are fused into a robot-centric 2.5D map
using \texttt{elevation\_mapping\_cupy} and its multi-modal
extension~\cite{elemap,semelemap}. The scores are temporally fused and
thresholded at $\tau=0.5$, and the resulting map is cropped to the same
yaw-aligned $41\times21$ terrain-affordance representation used in training.
The global map is accumulated at $3\,\mathrm{Hz}$, while the local
terrain-affordance crop is re-extracted at $20\,\mathrm{Hz}$ from the current
base pose.

\subsubsection{Real-World Results}
\label{sec:realworld_results}

All hardware trials use the same checkpoint, threshold $\tau$, and
fusion parameters, with no per-scene tuning or pre-mapping. Velocity commands
are provided by a wireless joystick.

The indoor experiments of Fig.~\ref{fig:indoor_pipes_grass} isolate semantic
contact selection under reliable segmentation. In the stepping-stone scene
(a), cardboard boxes and concrete rails have similar height and footprint,
but only the boxes are flagged. The robot traverses using the rails while
keeping its feet off the boxes, ruling out a general aversion to raised
geometry. It also clears a $30\,\mathrm{cm}$ pipe with both foot and shank
(b), avoids turf while moving forward, backward, and sideways (c), and
crosses three low pipes in both directions without contact (d). The turf
thickness lies within the map-height noise used during training, making the
cost channel the distinguishing cue from the surrounding floor.

The outdoor experiments use the same perception and control stack under
natural lighting and real materials. On the tiled walkway of
Fig.~\ref{fig:outdoor_lava_tiles}, narrow grass gaps leave only small
permitted support regions after max pooling. With segmentation active,
the robot crosses in both directions using the tiles; with segmentation
disabled, the same policy traverses the same geometry while stepping onto
the grass. This controlled comparison isolates the cost channel as
the cause of the change in foothold selection. The robot additionally clears
a narrow pipe on a timber deck and keeps its feet on pavement while walking
along an irregular grass boundary (Fig.~\ref{fig:outdoor_pipe_and_grass}),
demonstrating operation with thin obstacles and noisy natural boundaries.

Comparing ours to prior works, the semantic navigation
methods~\cite{wellhausen2019where,frey2023fast,kim2024semantic,roth2024viplanner}
act on the path: a flagged region is avoided by routing around it, and a
region that must be crossed, such as the pipes of
Fig.~\ref{fig:outdoor_pipe_and_grass}(a), is left to a geometric controller
that cannot tell a flagged support from a permitted one.
SemLoco~\cite{liang2026semloco} acts at the foothold level, but its
hazards all protrude from a flat floor, where its own ablation shows
geometry serving as a proxy for the hazard.

\section{Conclusion}

We presented SABER, a planner-free locomotion policy that jointly reasons about
terrain geometry and semantic contact permission through a unified
terrain-affordance map and a structured, foot-relative attention bias.
Simulation ablations and Unitree B2 experiments show that semantic information can directly 
shape contact selection: the policy avoids flagged terrain while still stepping on geometrically identical permitted supports.
These results suggest that contact permission can be integrated into the
low-level locomotion policy rather than handled only through higher-level
navigation or external foothold planning.

The main limitations arise from deployment perception and from the soft nature
of the learned constraint. Projecting 2D segmentation into a 2.5D elevation
map introduces sensitivity to alignment, map drift, and semantic prediction
errors, while sufficiently strong velocity commands can still drive the policy
through flagged terrain when no permitted path exists.
Future work will model semantic uncertainty during training, investigate
egocentric representations that avoid long-horizon map accumulation, and
extend semantic contact reasoning to other legged embodiments, including
humanoids.

\section*{Acknowledgment}
This research is supported by the Home Team Science and Technology Agency (HTX).
Claude Code (Anthropic) assisted with coding and debugging
(Secs.~\ref{sec:training}, \ref{sec:hardware_deployment}); Claude
(Anthropic) was used to improve the manuscript's clarity and consistency,
including rephrasing passages drafted by the authors. All content was
reviewed and verified by the authors, and all claims are their own.

\bibliographystyle{IEEEtran}
\bibliography{references}

\end{document}